\documentclass{article} 
\usepackage{iclr2026_conference,times}

\usepackage{amsmath,amsfonts,bm}

\def\eqref#1{equation~\ref{#1}}

\def\1{\bm{1}}

\def\va{{\bm{a}}}

\def\vg{{\bm{g}}}
\def\vh{{\bm{h}}}

\def\vr{{\bm{r}}}

\def\vx{{\bm{x}}}

\def\vz{{\bm{z}}}

\DeclareMathAlphabet{\mathsfit}{\encodingdefault}{\sfdefault}{m}{sl}
\SetMathAlphabet{\mathsfit}{bold}{\encodingdefault}{\sfdefault}{bx}{n}

\usepackage{hyperref}
\usepackage{url}
\usepackage{booktabs}
\usepackage{multirow}
\usepackage{graphicx}
\usepackage{subcaption}
\usepackage{float}

\title{When Object-Centric World Models Meet Policy Learning: From Pixels to Policies, and Where It Breaks}

\author{
Stefano Ferraro$^{1,2}$ \quad Akihiro Nakano$^{2}$ \quad Masahiro Suzuki$^{2}$ \quad Yutaka Matsuo$^{2}$ \\
$^{1}$IDLAB, Gent University, Gent, Belgium \\
$^{2}$Graduate School of Engineering, The University of Tokyo, Tokyo, Japan \\
Corresponding Email: \texttt{stefano.ferraro@ugent.be}
}
\iclrfinalcopy 
\begin{document}

\maketitle

\begin{abstract}
Object-centric world models (OCWM) aim to decompose visual scenes into object-level representations, providing structured abstractions that could improve compositional generalization and data efficiency in reinforcement learning. We hypothesize that explicitly disentangled object-level representations, by localizing task-relevant information, can enhance policy performance across novel feature combinations. To test this hypothesis, we introduce DLPWM, a fully unsupervised, disentangled object-centric world model that learns object-level latents directly from pixels. DLPWM achieves strong reconstruction and prediction performance, including robustness to several out-of-distribution (OOD) visual variations. However, when used for downstream model-based control, policies trained on DLPWM latents underperform compared to DreamerV3. Through latent-trajectory analyses, we identify representation shift during multi-object interactions as a key driver of unstable policy learning. Our results suggest that, although object-centric perception supports robust visual modeling, achieving stable control requires mitigating latent drift.  

\end{abstract}

\section{Introduction}



An enduring challenge in artificial intelligence is the development of agents that can reason about the world in a structured, human-like manner. A promising avenue towards this goal lies in the use of object-centric world models (OCWMs), which aim to decompose complex visual scenes into object-level representations~\citep{greff2020binding}. Such structured abstractions have shown significant potential for improving compositional generalization and data efficiency in various downstream tasks. Seminal works in this area have demonstrated that object-centric representations, often learned in an unsupervised fashion, can facilitate robust object property prediction and exhibit resilience to certain out-of-distribution (OOD) visual variations~\citep{dittadi2022generalization,locatello2020slotattn,kipf2022conditional}. Furthermore, recent studies suggest that these representations can lead to better compositional generalization with reduced computational overhead~\citep{kapl2025object}.

Building on these findings, a key question arises: can structured, object-centric representations improve policy learning in reinforcement learning (RL)?
Although recent studies show gains in relational or compositional tasks~\citep{mosbach2025sold,zhang2025objects,yoon2023investigation}, these often fail to generalize beyond the training distribution.
We posit that this limitation arises from a misalignment between representation and control, where object-centric models still entangle task-relevant and irrelevant factors—yielding policies that are structured but brittle.

We hypothesize that the key to unlocking the full potential of object-centric representations for policy generalization lies in the disentanglement of task-relevant features. By isolating the factors of variation that are crucial for a given task, the resulting latent space becomes a more reliable and robust foundation for the policy, particularly in novel or out-of-distribution scenarios. A policy that can consistently access and reason about these disentangled features should, in principle, generalize more effectively to unseen combinations of object properties and environmental conditions.

Disentangled object-centric methods pursue this goal by producing per-object latents that separate spatial attributes (e.g., position, scale) from appearance and dynamics~\citep{jiang2020scalor,lin2020improving,nakano2023interactionbased}. For example, Deep Latent Particles (DLP) represent objects as learned particles with explicit spatial and feature components, enabling interpretable, unsupervised learning from pixels~\citep{daniel2022dlp}. Its temporal extension, Deep Dynamic Latent Particles (DDLP), augments DLP with dynamics-aware particles for object-level video prediction~\citep{daniel2024ddlp}.

To test our hypothesis, we introduce DLPWM, a fully unsupervised, disentangled object-centric world model that learns object-level latents directly from pixels. In our initial evaluations, DLPWM demonstrates strong performance in visual modeling, achieving high-fidelity reconstruction and accurate prediction, even in the presence of several OOD visual variations. This suggests that the model successfully learns a robust and generalizable representation of the visual scene. However, when these learned representations are leveraged for downstream model-based control, policies trained on DLPWM latents underperform in comparison to the state-of-the-art holistic world model, DreamerV3~\citep{hafner2023dreamer3}. Through in-depth latent-trajectory analyses, we identify a significant ``representation shift'' during multi-object interactions as a primary driver of this unstable policy learning.

\section{Related Works}
\paragraph{Model-based RL.}
Model-based reinforcement learning (MBRL) improves efficiency and generalization by learning environment dynamics for prediction, planning, and imagination-based policy training. Early neural world models, such as PlaNet~\citep{hafner2019planet} and the Dreamer series~\citep{hafner2019dreamer1,hafner2020dreamer2,hafner2023dreamer3}, showed that latent dynamics can replace explicit simulators, enabling strong pixel-based control with far fewer samples than model-free approaches. More recent work enhances temporal expressiveness by leveraging Transformers to capture long-term dependencies and stabilize imagination rollouts~\citep{micheli2023transformers,robine2023transformerbased,chen2022transdreamer,zhang2023storm,meo2025masked,nakano2023interactionbased}, establishing latent world models as a cornerstone of visual RL.

\paragraph{Object-centric World Models.}
Conventional MBRL represents latent states as unstructured vectors, whereas object-centric approaches impose compositional structure by decomposing scenes into entities or “slots” with disentangled attributes~\citep{burgess2019monet,greff2019multi,locatello2020object}. Such structured representations improve interpretability and relational reasoning~\citep{veerapaneni2019op3,greff2020binding,wu2023slotformer}. Building on this, Object-Centric World Models (OCWMs)~\citep{mosbach2025sold,ferraro2023focus,zhang2025objects,nishimoto2024transformerbased} integrate slot-based perception with latent dynamics, enabling agents to learn multi-object interactions directly from pixels. Empirical results from SOLD~\citep{mosbach2025sold} and FOCUS~\citep{ferraro2023focus,ferraro2024representing} show benefits on relational control tasks, though broader evidence across diverse benchmarks remains limited.

\section{Method}

\subsection{Preliminary: Dynamic Deep Latent Particles}
We build upon the Dynamic Deep Latent Particles (DDLP) architecture~\citep{daniel2022dlp, daniel2024ddlp}, a method for unsupervised representation learning that disentangles object position from appearance. DDLP decomposes a visual input into a set of low-dimensional latent ``particles'', where each particle is described by its spatial location, depth, scale, transparency and a visual feature vector. This VAE-based approach provides a structured and interpretable representation of a scene.

Formally, for a given observation $\vx_t$, the model processes it as follows:
\begin{equation}
\begin{split}
\text{Encoder:} \quad & \vz_t^{1:K} \sim e_{\psi}(\vz_t^{1:K} | \vx_t), \\
\text{Decoder:} \quad & \hat{\vx}_t \sim d_{\psi}(\hat{\vx}_t | \vz_t^{1:K}), \\
\text{Dynamics predictor:} \quad & \vh_{1:t+1}^{0:K} = p_{\psi}(\vz_{0:t}^{0:K}), \\
\text{Particle decoder:} \quad & \hat{\vz}_t^{0:K} \sim d_{\psi}(\hat{\vz}_t^{0:K} | \vh_{t}^{0:K}),
\end{split}
\end{equation}
where $\{\vz_t^k\}_{k=1}^K$ denotes the set of $K$ latent particles at time $t$. Each particle are expressed as $\vz_t^k = (\vz_{p,t}^k, \vz_{d,t}^k, \vz_{s,t}^k, \vz_{\tau,t}^k, \vz_{f,t}^k)$, where $\vz_{p,t}^k$ is the \textit{pixel position}, $\vz_{d,t}^k$ represents the \textit{depth information}, $\vz_{s,t}^k$ is the \textit{entity scaling}, $\vz_{\tau,t}^k$ is the \textit{transparency feature} and $\vz_{f,t}^k$ is the \textit{visual features latent}. The functions $e_{\psi}$ and $d_{\psi}$ represent the encoder and decoder, respectively. 

A dynamics model, $p_{\psi}$, is trained to predict the future states given the past latent trajectory. The model is trained by optimizing a modified evidence lower bound (ELBO) which is inspired by the Chamfer distance between particles. 

\subsection{DLPWM}

To construct a world model from DDLP, we first condition the dynamics predictor on the agent’s actions $\va_{0:t}$. Unlike Dreamer, our model operates on a structured latent space, where information for each scene entity is disentangled. Many tasks require relating information across entities (e.g., reaching an object requires combining positional data from both agent and object). To capture such relations for accurate reward prediction, we use a particle aggregator that takes $\vh_{t}^{0:K}$ as input and outputs $\vg_{t}$. Next, to enable training of a joint policy in imagination, we add a reward prediction head. The aggregated information $\vg_t$ is provided as input to the reward predictor. The reward predictor is trained by minimizing the mean squared error between predicted and ground-truth rewards:
$\mathcal{L}_{\mathrm{reward}} \;=\; \mathbb{E}_{t}\big[(\hat{\vr}_{t} - \vr_{t})^{2}\big]$. 

\begin{equation}
\begin{split}
\text{Dynamics predictor:} \quad & \vh_{1:t+1}^{0:K} = p_{\psi}(\vz_{0:t}^{0:K}, \va_{0:t}), \\
\text{Particle aggregator:} \quad & \vg_{t} = p_{\psi}(\vh_{t}^{0:K}), \\
\text{Reward predictor:} \quad & \hat{\vr}_{t} \sim p_{\psi}(\hat{\vr}_{t} | \vg_{t}),
\end{split}
\end{equation}

Finally, we introduce a policy module based on the actor–critic architecture of DreamerV3~\citep{hafner2023dreamer3}. Similar to the reward predictor, the policy components use a particle aggregator to relate information across particles. Each of the reward predictor, actor, and critic employs its own independent aggregator module.

\begin{equation}
\text{Actor: }  \va_{t} \sim \pi_{\theta}(\va_{t} | \vg_{t}), \quad \text{Critic: } v_{\gamma}(R_t | \vg_{t}) \\
\end{equation}

\begin{table}[b]
\centering
\scriptsize
\setlength{\tabcolsep}{4pt}
\renewcommand{\arraystretch}{1.1}
\begin{tabular}{l l c c | c c} 
 & & \multicolumn{2}{c|}{\textbf{Reconstruction}} & \multicolumn{2}{c}{\textbf{Prediction}} \\
\cmidrule(lr){2-4} \cmidrule(lr){5-6} 
 & & SSIM $\Uparrow$ & LPIPS $\Downarrow$ & SSIM $\Uparrow$ & LPIPS $\Downarrow$ \\
\midrule
\multirow{2}{*}{Cube lifting} 
  & DreamerV3 & $0.979 \pm 0.001$ & $0.05 \pm 0.0012$ & $0.89 \pm 0.019$ & $0.08 \pm 0.01$ \\
  & DLPWM (ours) & $0.976 \pm 0.003$ & \textbf{0.0384} $\pm$ \textbf{0.0025} & $0.9 \pm 0.02$ & $0.08 \pm 0.019$ \\
\midrule
\multirow{2}{*}{Generalization Arena ID} 
  & DreamerV3 & $0.933 \pm 0.012$ &  $0.102 \pm 0.012$ & \textbf{0.833} $\pm$ \textbf{0.033} & $0.122 \pm 0.016$  \\
  & DLPWM (ours) &  $0.943 \pm 0.012$ & \textbf{0.066} $\pm$ \textbf{0.014} & $0.799 \pm 0.031$  & $ 0.11 \pm 0.015$  \\
\midrule
\multirow{2}{*}{Generalization Arena OOD} 
  & DreamerV3 & $0.936 \pm 0.01$ & $0.096 \pm 0.007$ & \textbf{0.837} $\pm$ \textbf{0.033} & $0.118 \pm 0.014$ \\
  & DLPWM (ours) & \textbf{0.943} $\pm$ \textbf{0.011} & \textbf{0.064} $\pm$ \textbf{0.01} & $0.792 \pm 0.037$ & $0.118 \pm 0.015$\\
\bottomrule
\end{tabular}
\caption{Reconstruction and prediction metrics for and DLPWM. Both world models are trained for 50k using an offline dataset (for 2 seeds). Predictions are done over 15 steps. Metrics are averaged over 10 evaluation episodes. $\pm$ confidence intervals.}
\label{tab:recon_pred}
\end{table}

\section{Experiments}
We benchmark our model against DreamerV3~\citep{hafner2023dreamer3}. Both world models are trained offline on a standard block-lifting task and a generalized object-lifting task. For the latter, we introduce a novel Robosuite environment~\citep{Zhu2020Robosuite}, the Generalization Arena (see Appendix~\ref{sec:gen_arena}), which allows control over object shapes and colors. To test the generalization capabilities of our model, during training only a subset of shape–color combinations is available, enabling evaluation on out-of-distribution configurations.

We first compare reconstruction and prediction performance between models. Next, we analyze the learned policies to highlight the impact of object-centric learning. Finally, we discuss potential factors contributing to our model’s underperformance.

\paragraph{Reconstructions and Predictions.}
We train both DLPWM and the DreamerV3 baseline on the cube-lift and Generalization Arena tasks. Evaluation results are reported in Table~\ref{tab:recon_pred}, considering both in-distribution (ID) and out-of-distribution (OOD) cases for the Generalization Arena. DLPWM generally achieves better reconstruction performance, while prediction accuracy remains comparable between the two models.

\paragraph{Policy Learning.}
After 50k steps of offline DLPWM training, online policy learning is initiated concurrently to the world model training. We evaluate two implementations of the particle aggregator: GNN-based and Transformer-based. Results are shown in Figure~\ref{fig:policy_learning}. Both policies successfully learn to reach the target object (initial 100k steps) but struggle with grasping and lifting. We hypothesize that this limitation arises from direct physical interactions between entities—when objects make contact, the particle representation may fail to preserve clear boundaries between them.

\begin{minipage}[t]{0.48\linewidth} 
    \centering
    \begin{figure}[H]
    \centering
    \includegraphics[width=0.7\linewidth]{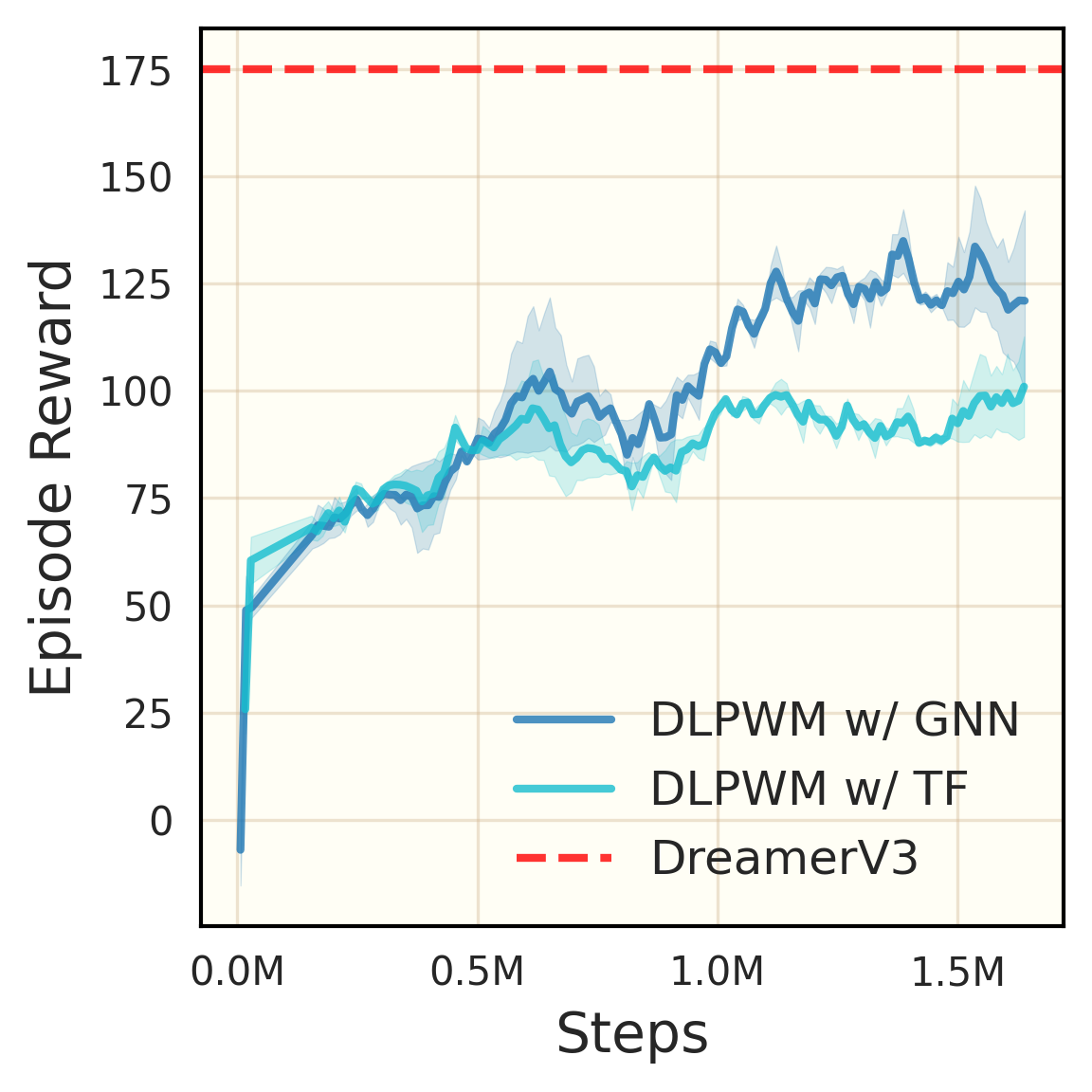}
    \caption{Per episode reward over training steps. Trained on cube lift task. Policy trained with DLPWM, are tested with both GNN and transformer (TF) particle aggregator. During the policy training phase, world model and policy update occur every 10 steps. 2 seed for each run are considered.}
    \label{fig:policy_learning}
    \end{figure}
\end{minipage}
\hfill 
\begin{minipage}[t]{0.48\linewidth} 
    \centering
        \begin{figure}[H]
        \centering
        \includegraphics[width=0.7\linewidth]{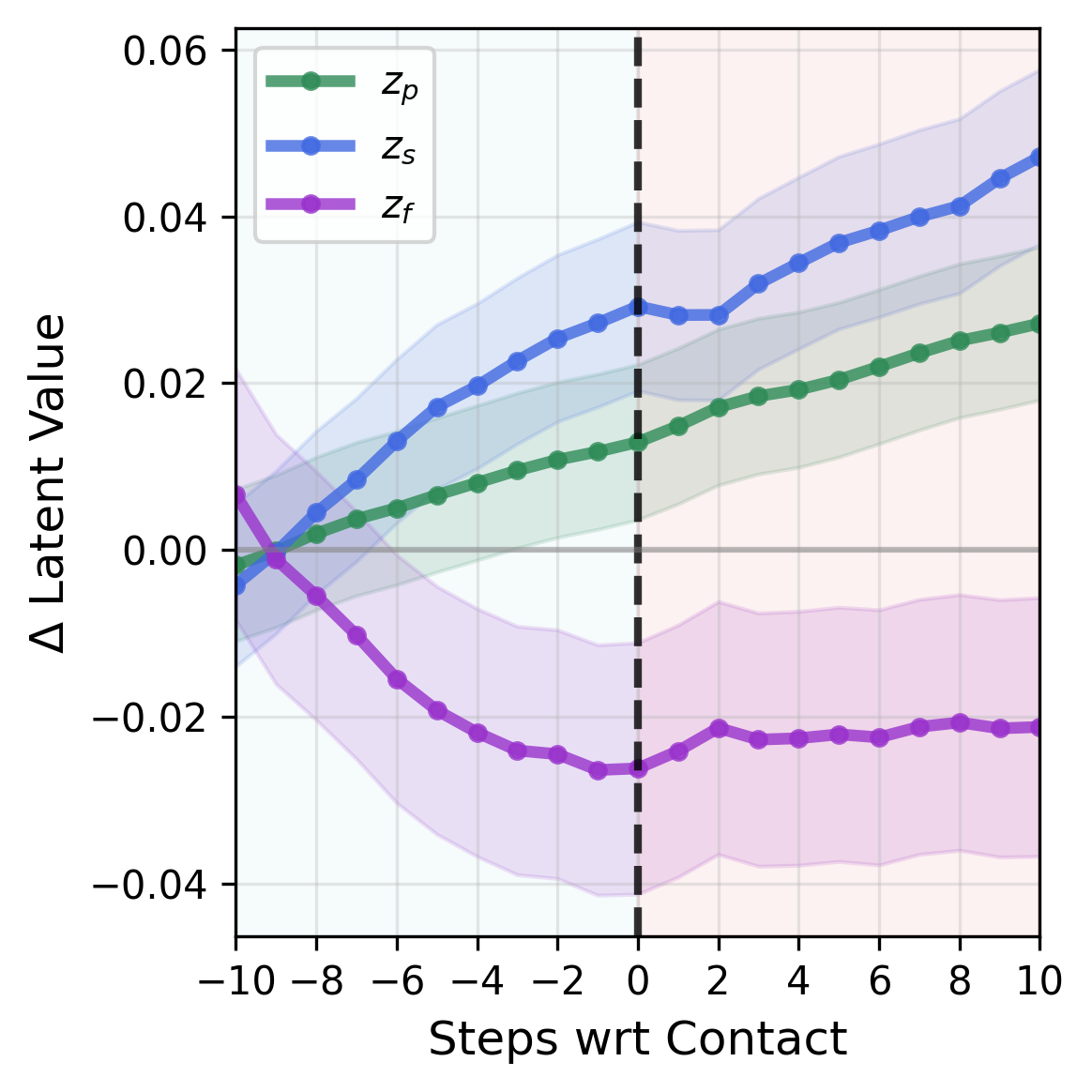}
        \caption{Latent variation with respect to contact point frames. The horizontal dashed-line represent the frame where contact between the robotic arm and the target object is established. Visualized are the position latent $z_p$, scale latent $z_s$ and the visual features latent $z_f$. Results are averaged over 10 evaluation episodes where a total of 39 contact points are identified.}
        \label{fig:contact_latents}
        \end{figure}
\end{minipage}

\paragraph{Latent Trajectories.}
We analyze the latent representations of each object during contact events, as shown in Figure~\ref{fig:contact_latents}. Near contact frames, the positional ($z_p$), scale ($z_s$), and visual feature ($z_f$) components of the latent space exhibit noticeable changes. Such alterations should occur only when the object interacts physically with the robotic arm (hence at frame 0 to 10); however, the arm’s proximity alone can perturb the object’s latent features (from frames -10 to 0). We hypothesize that this interference contributes to the observed policy underperformance. A visual example of this behavior is provided in Figure~\ref{fig:reconstructions}.a in the appendix, near the grasping frame (Frame 9).

\vspace{-0.3cm}
\section{Conclusion and Discussion}


We introduce DLPWM, a disentangled object-centric world model that attains strong reconstruction and prediction but whose imagined latents sometimes yield unstable policy learning on interaction-heavy task phases.
We hypothesize that brief, contact-induced perturbations (and occasional slot-identity drift) corrupt the per-slot inputs the policy uses.

As next steps, we propose to supply the policy with an exponential moving average (EMA) of each slot — concatenated with the raw slot (and optionally the slot-delta) — which (i) attenuates high-frequency corruption, (ii) is trivial to compute online in both real and imagined rollouts, and (iii) preserves fast signals when raw+delta are also provided.

\section*{Acknowledgments}
Stefano Ferraro was supported by the Research Foundation Flanders (FWO), grant number V431625N, and by the Flemish Government through the AI Research Program.

\bibliography{iclr2026_conference}
\bibliographystyle{iclr2026_conference}

\newpage
\appendix
\section{Appendix}
\subsection{Generalization Arena}
\label{sec:gen_arena}
We extend the standard Cube Lift RoboSuite environment to support multiple object types. Object properties are parameterized by shape and color, with available shapes \texttt{[cube, ring, L-profile]} and colors \texttt{[red, blue, green]}. At each episode, an object combination is randomly selected from a predefined set.

The environment provides a stepwise reward signal identical to the Cube Lift task, structured into three phases: reaching, grasping, and lifting. The maximum reward is achieved when the agent successfully grasps and lifts the object from the table.

\begin{figure}[t]
\centering
\includegraphics[width=\linewidth]{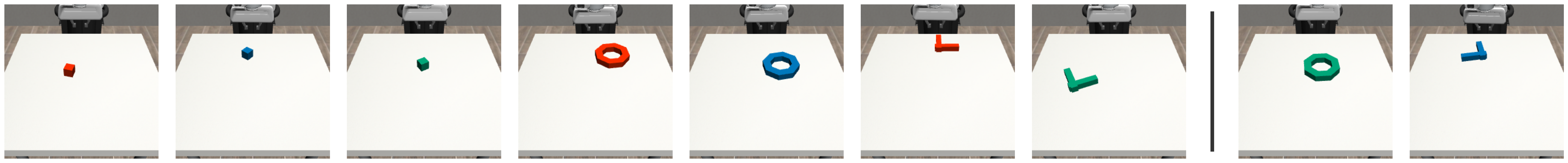}
\caption{All the shape and color combinations present in the Generalization Arena task. On the left, 7 combinations used for training and one the right 2 combinations used for evaluation.}
\label{fig:env_combs}
\end{figure}

\subsection{DLPWM}

\begin{figure}[H]
\centering
\includegraphics[width=0.9\linewidth]{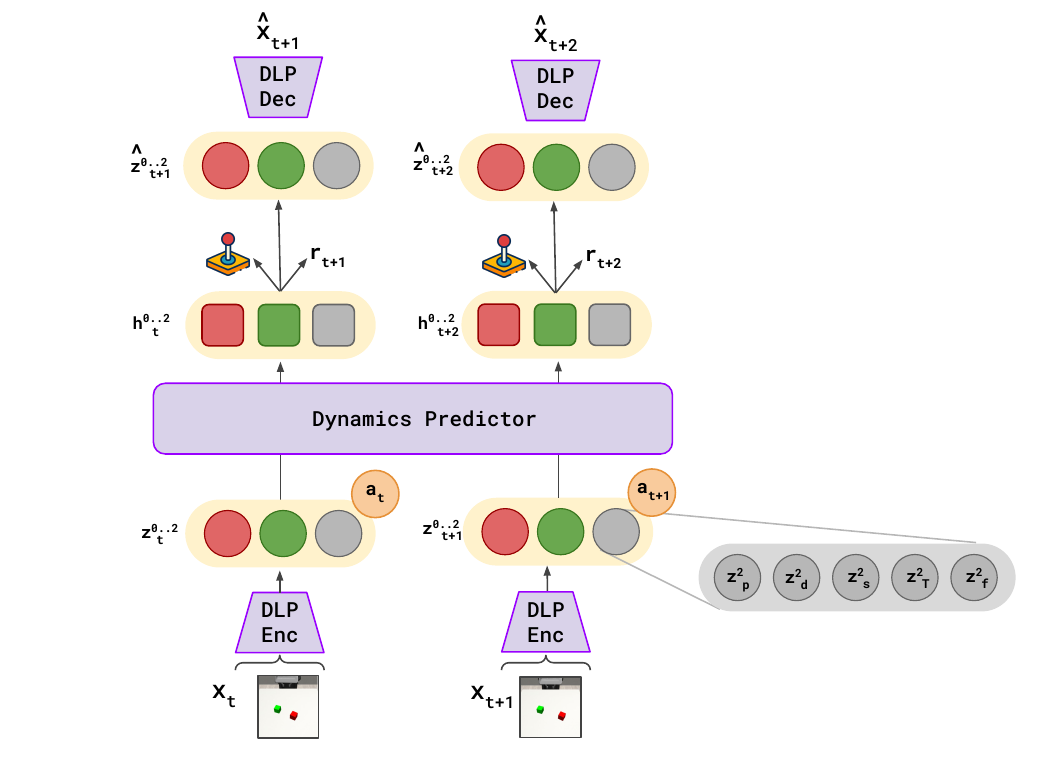}
\caption{Architecture of DLPWM.}
\label{fig:dlpwm}
\end{figure}

\subsection{Particle aggregator}
For our experimentation we tested 2 implementation for the particle aggregator. The first uses a GraphSAGE-based GNN, where each particle corresponds to a node and an additional fully connected output node aggregates information from all others.
The second follows a Transformer-based design inspired by the Slot Attention Transformer (SAT) from SOLD~\cite{mosbach2025sold}, using one register token and one output token, with only the latter used as the final output. 

\newpage
\subsection{Reconstructions}
\begin{figure}[h!]
    \centering 

    \begin{subfigure}[t]{0.9\linewidth}
        \centering
        \includegraphics[width=\linewidth]{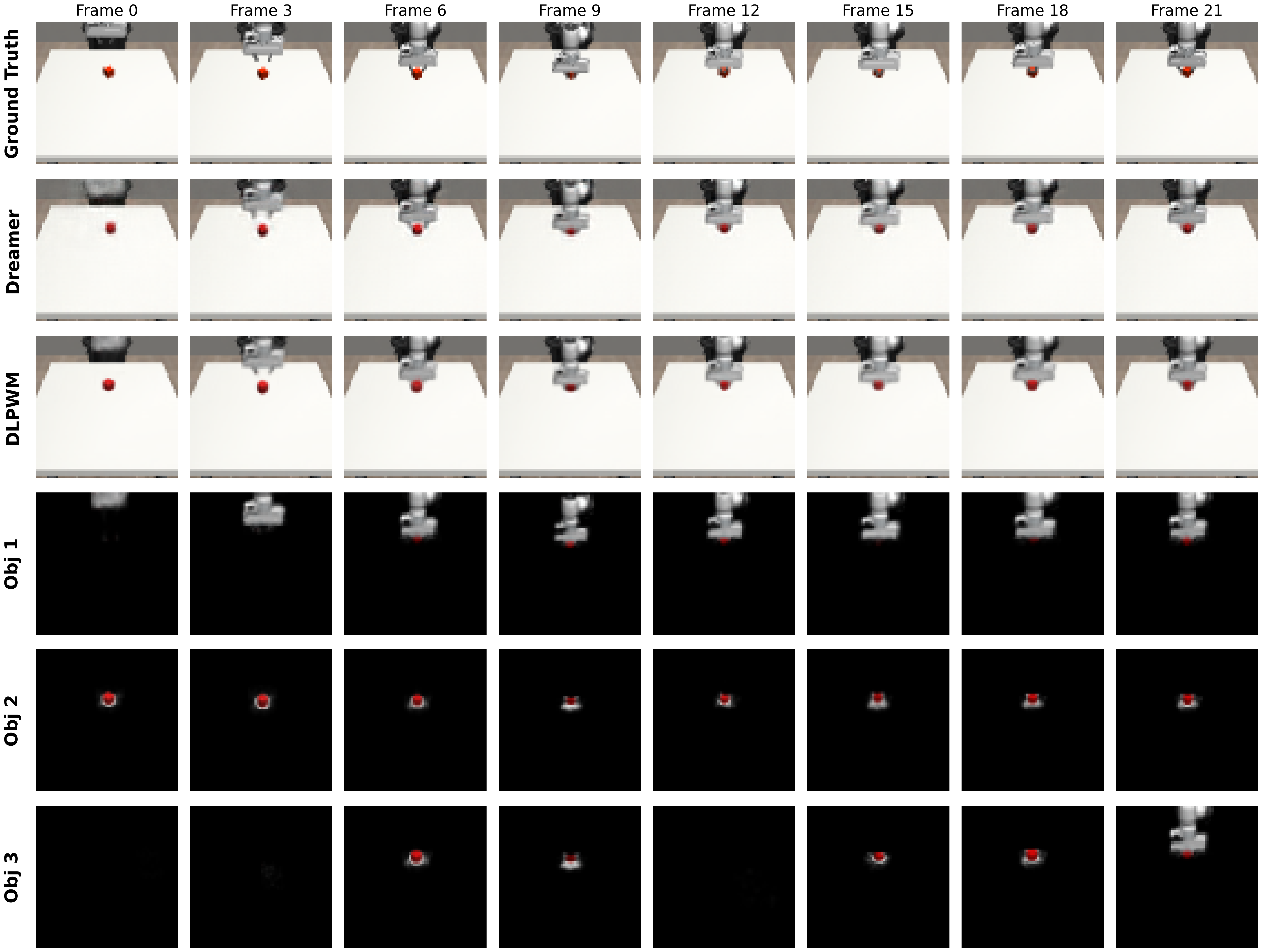}
        \caption{Cube task.}
        \label{fig:a_rec}
    \end{subfigure}
    \vfill 
    
    \begin{subfigure}[t]{0.9\linewidth}
        \centering
        \includegraphics[width=\linewidth]{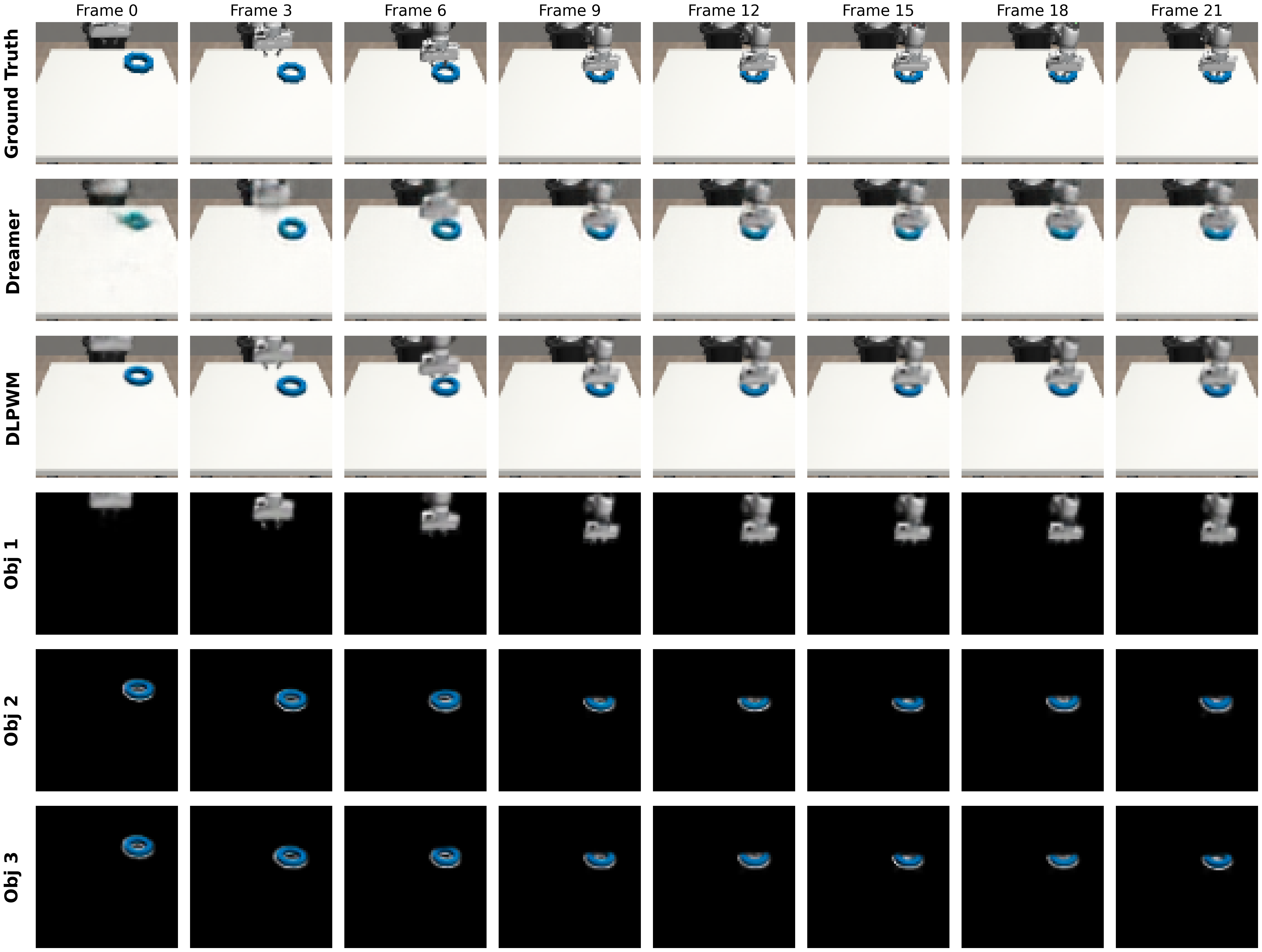}
        \caption{Generalization Arena in-distribution.}
        \label{fig:b_rec}
    \end{subfigure}
\end{figure}

\begin{figure}[t]\ContinuedFloat
    \centering 
    \setcounter{subfigure}{2} 
    \begin{subfigure}[t]{0.9\linewidth}
        \centering
        \includegraphics[width=\linewidth]{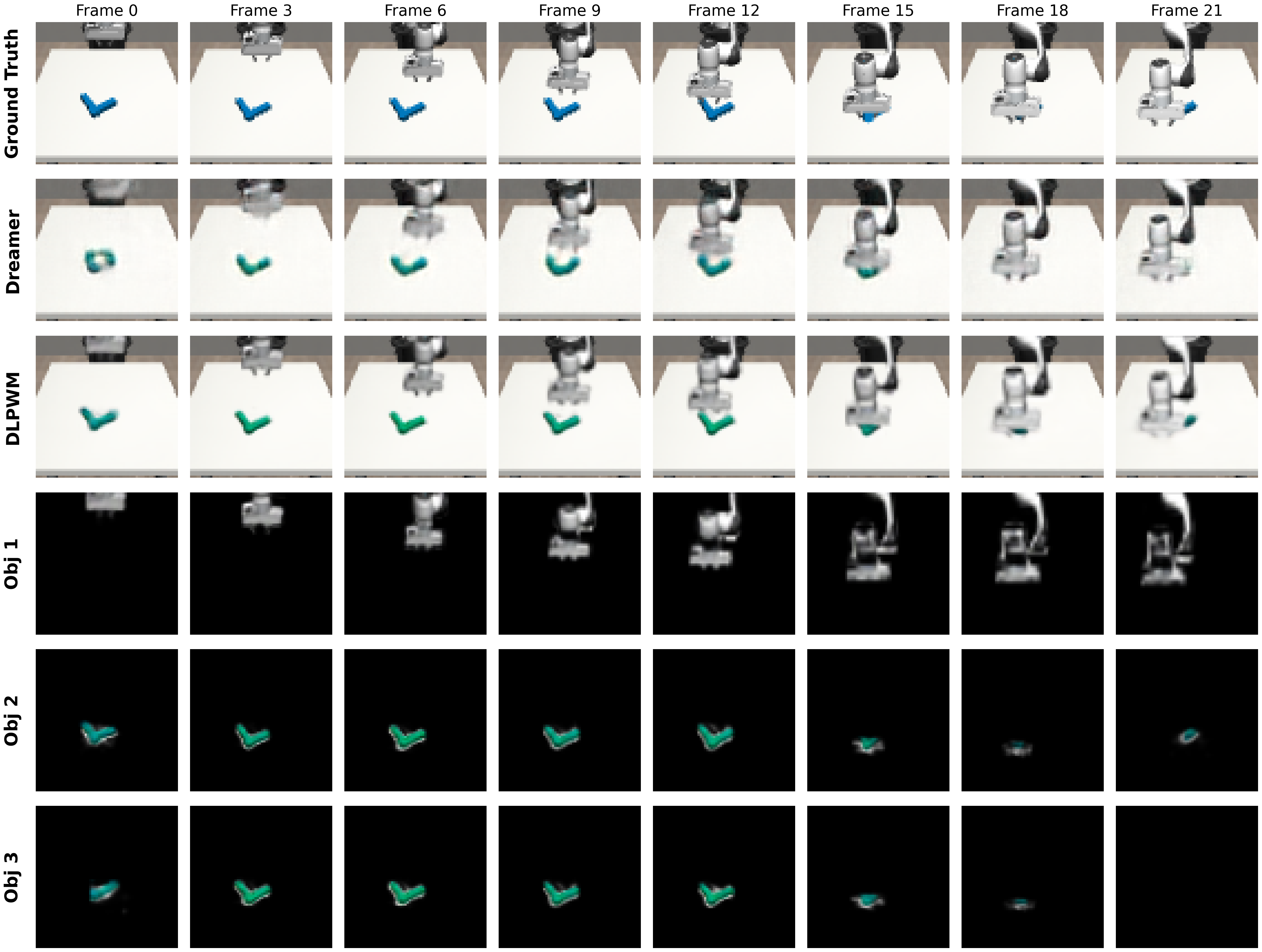}
        \caption{Generalization Arena out-of-distribution.}
        \label{fig:c_rec}
    \end{subfigure}

    \caption{Reconstruction examples for DreamerV3 and DLPWM. Object masks are from DLPWM.}
    \label{fig:reconstructions}
\end{figure}

\newpage
\subsection{Predictions}

\begin{figure}[!b]
    \centering 

    \begin{subfigure}[t]{0.9\linewidth}
        \centering
        \includegraphics[width=\linewidth]{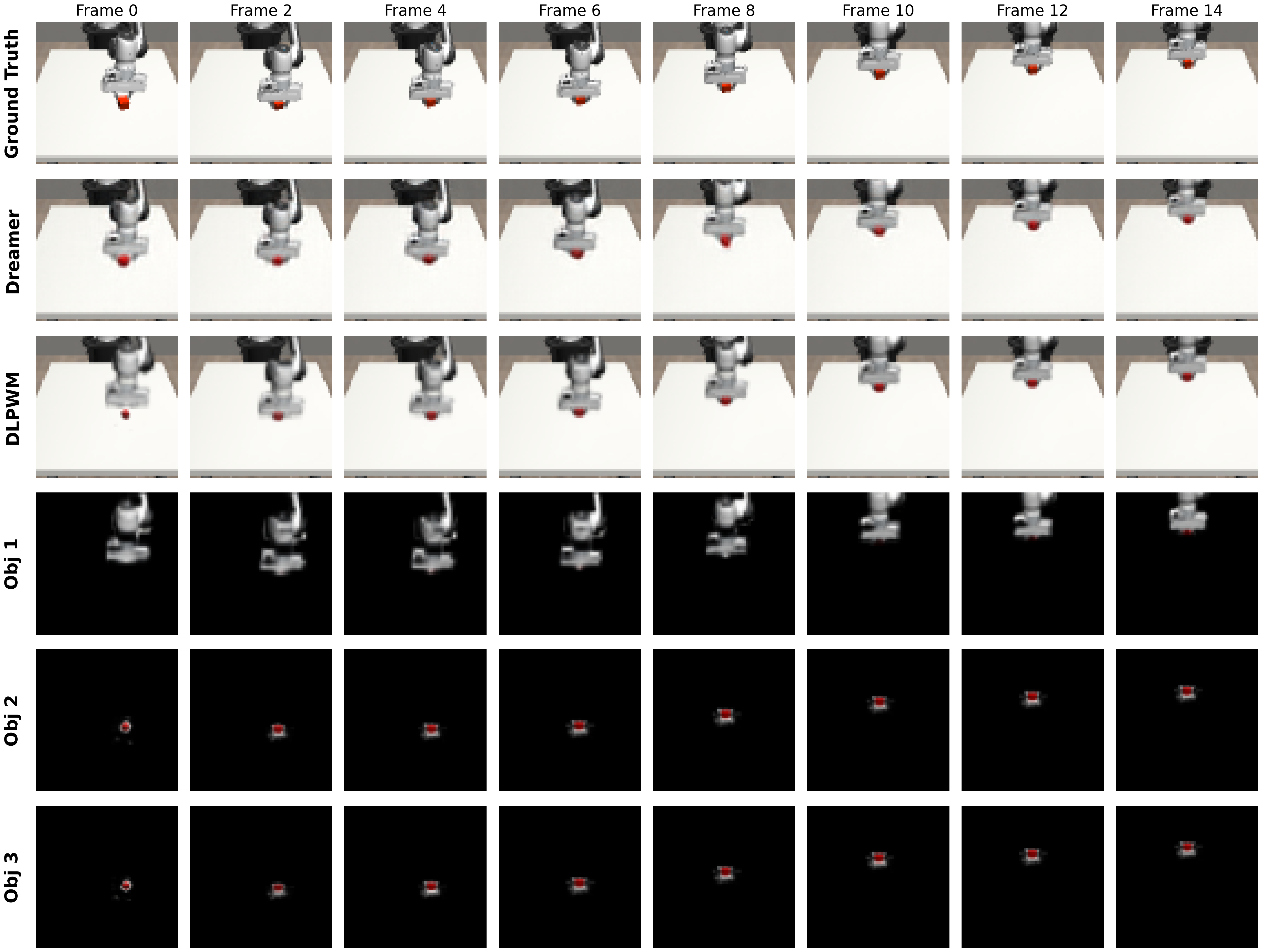}
        \caption{Cube task.}
        \label{fig:a_pred}
    \end{subfigure}
    \vfill 
\end{figure}
\begin{figure}[t]\ContinuedFloat
    \centering
    \setcounter{subfigure}{1} 
    \begin{subfigure}[t]{0.9\linewidth}
        \centering
        \includegraphics[width=\linewidth]{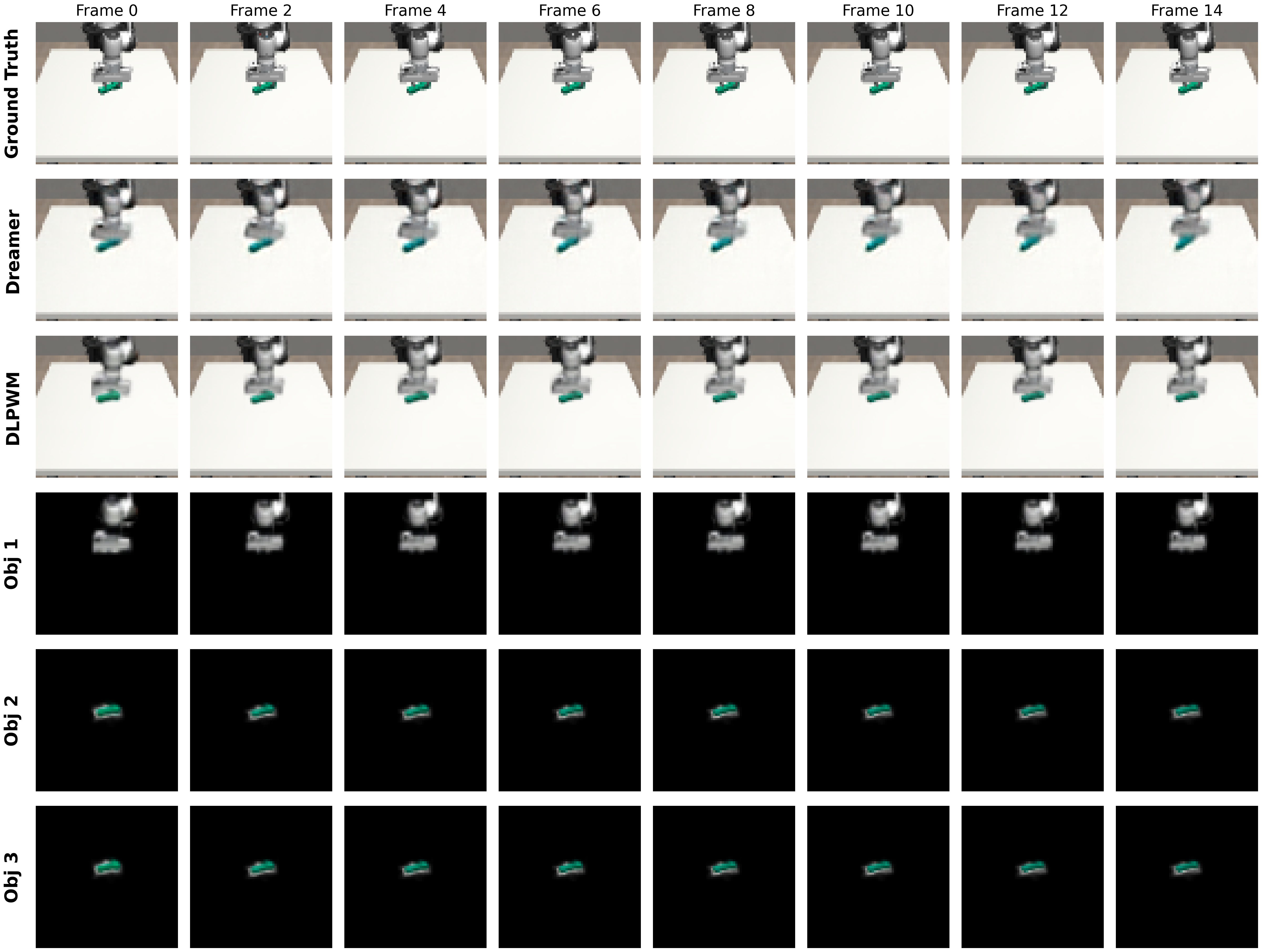}
        \caption{Generalization Arena in-distribution.}
        \label{fig:b_pred}
    \end{subfigure}
    \vfill
    \begin{subfigure}[t]{0.9\linewidth}
        \centering
        \includegraphics[width=\linewidth]{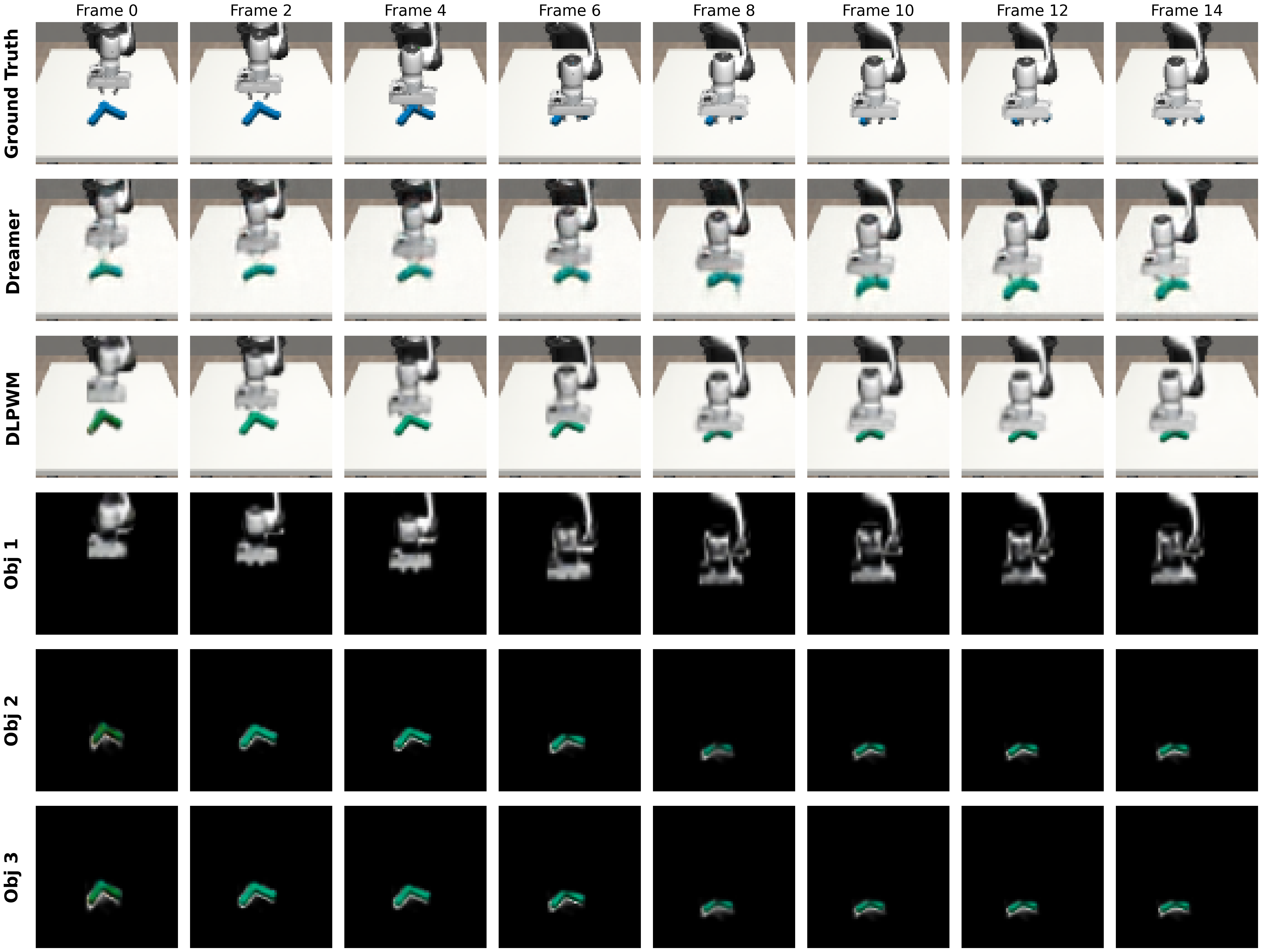}
        \caption{Generalization Arena out-of-distribution.}
        \label{fig:c_pred}
    \end{subfigure}

    \caption{Prediction examples for DreamerV3 and DLPWM. Object masks are from DLPWM.}
    \label{fig:predictions}
\end{figure}

\end{document}